\documentclass[10pt,twocolumn,letterpaper]{article}

\usepackage{iccv}
\usepackage{times}
\usepackage{epsfig}
\usepackage{graphicx}
\usepackage{amsmath}
\usepackage{amssymb}

\usepackage{ragged2e}
\usepackage{booktabs}
\usepackage{multirow}
\usepackage{makecell}
\usepackage[accsupp]{axessibility}
\usepackage{appendix}
\usepackage{float}

\usepackage[breaklinks=true,bookmarks=false]{hyperref}

\iccvfinalcopy 

\def\iccvPaperID{4834} 

\ificcvfinal\fi

\begin{document}

\title{The Victim and The Beneficiary: Exploiting a Poisoned Model to Train a Clean Model on Poisoned Data}

\author{Zixuan Zhu, Rui Wang\thanks{Corresponding author} , 
 Cong Zou, Lihua Jing\\
SKLOIS, Institute of Information Engineering, CAS, Beijing, China\\
School of Cyber Security, University of Chinese Academy of Sciences, Beijing, China\\
{\tt\small \{zhuzixuan, wangrui, zoucong, jinglihua\}@iie.ac.cn}
}

\maketitle
\ificcvfinal\thispagestyle{empty}\fi


\begin{abstract}
    Recently, backdoor attacks have posed a serious security threat to the training process of deep neural networks (DNNs). The attacked model behaves normally on benign samples but outputs a specific result when the trigger is present. However, compared with the rocketing progress of backdoor attacks, existing defenses are difficult to deal with these threats effectively or require benign samples to work, which may be unavailable in real scenarios. In this paper, we find that the poisoned samples and benign samples can be distinguished with prediction entropy. This inspires us to propose a novel dual-network training framework: The Victim and The Beneficiary (V\&B), which exploits a poisoned model to train a clean model without extra benign samples. Firstly, we sacrifice the Victim network to be a powerful poisoned sample detector by training on suspicious samples. Secondly, we train the Beneficiary network on the credible samples selected by the Victim to inhibit backdoor injection. Thirdly, a semi-supervised suppression strategy is adopted for erasing potential backdoors and improving model performance. Furthermore, to better inhibit missed poisoned samples, we propose a strong data augmentation method, AttentionMix, which works well with our proposed V\&B framework. Extensive experiments on two widely used datasets against 6 state-of-the-art attacks demonstrate that our framework is effective in preventing backdoor injection and robust to various attacks while maintaining the performance on benign samples. 
    Our code is available at \href{https://github.com/Zixuan-Zhu/VaB}{https://github.com/Zixuan-Zhu/VaB}.
\end{abstract}
\vspace{-0.4cm}

\begin{figure}[t]
\begin{center}
\includegraphics[width=0.80\linewidth]{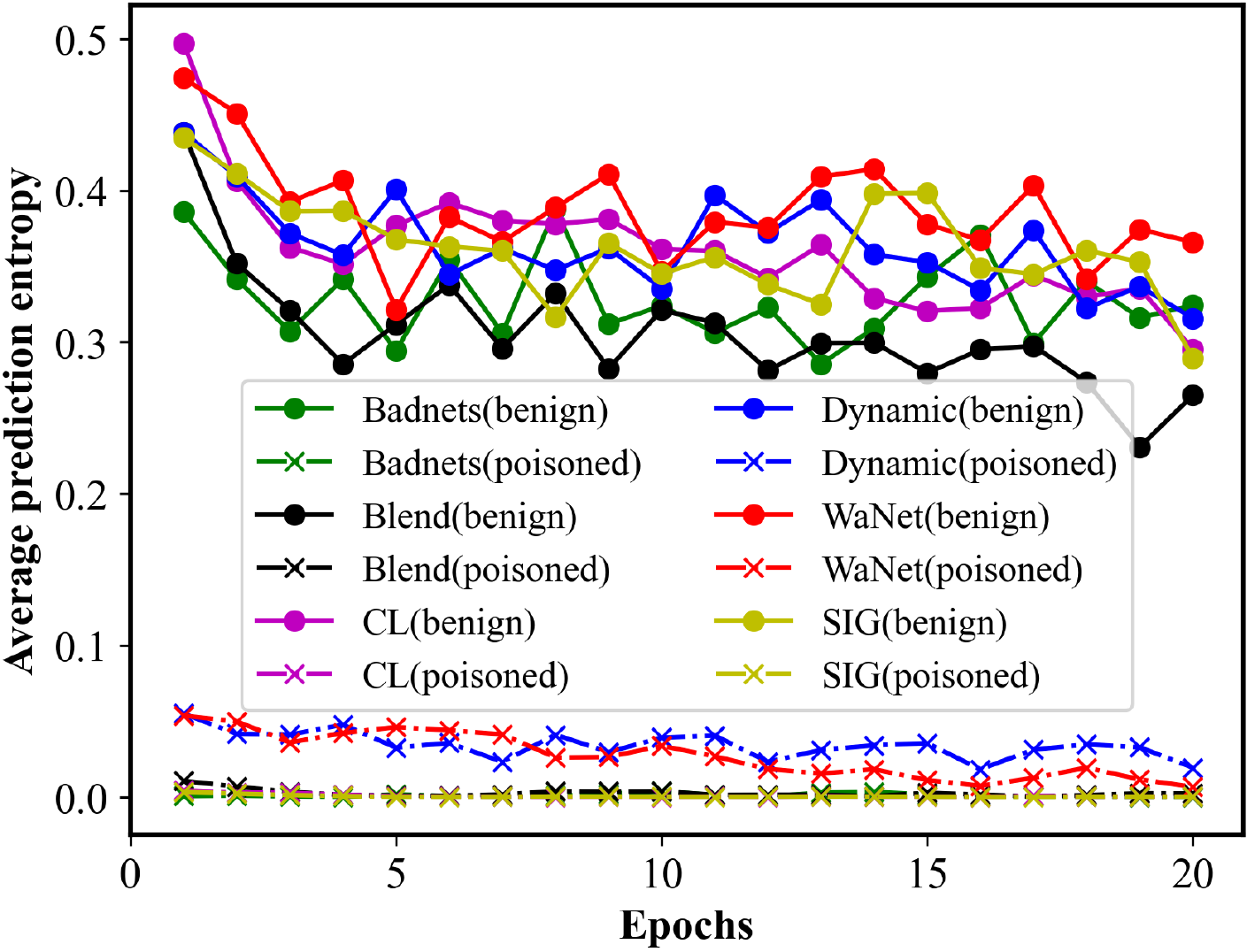}
\vspace{-0.2cm}
\end{center}
   \caption{The average prediction entropy of benign samples versus poisoned samples crafted by 6 backdoor attacks. We conduct the experiment on CIFAR-10 with ResNet-18 under poisoning rate 10\%, where our special training strategy (described in section \ref{section: warmup}) was adopted.}
\vspace{-0.5cm}
\label{fig: entropy}
\end{figure}

\section{Introduction}

Large amounts of training data and powerful computing resources are two key factors for the success of deep neural networks (DNNs). While in practical applications, obtaining enough samples for training is laborious, and training is likely to be resource-constrained. As a result, more and more third-party data (\eg\ data released on the Internet) and training platforms (\eg\ Google Colab) are adopted to reduce the overhead, which brings new security risks.

Backdoor attacks \cite{proflip, trojannet, qi2022towards} pose a serious security threat to the training process of DNNs and can be easily executed through data poisoning. Specifically, attackers \cite{blend, badnets, wanet} inject the designed \emph{trigger} (\eg\ a small patch or random noise) to a few benign samples selected from the training set and change their labels to the attacker-defined \emph{target label}. These poisoned samples will force models to learn the correlation between the trigger and the target label. In the reference process, the attacked model will behave normally on benign samples but output the target label when the trigger is present. Deploying attacked models can lead to severe consequences, even life-threatening in some scenarios (\eg\ autonomous driving). Hence, a secure training framework is needed when using third-party data or training platforms.

Many researchers have devoted themselves to this topic and proposed feasible defenses. Li \etal~\cite{NAD} utilize local benign samples to erase existing backdoors in DNNs, and Borgnia \etal~\cite{borgnia2021strong} employ strong data augmentations to inhibit backdoor injection during training. Li \etal~\cite{ABL} find that the training loss decreased faster for poisoned samples than benign samples and designed a loss function to separate them. Finally, they select a fixed portion of samples with the lowest loss to erase potential backdoors. In real scenarios, it is challenging to distinguish poisoned samples from benign ones in this way due to the unknown poisoning rate and close loss values. In experiments, we find the entropy of model prediction is a more discriminative property, and a fixed threshold could filter out most poisoned samples, as shown in Figure \ref{fig: entropy}. During training, poisoned samples will be learned faster than benign samples due to the similar feature of their triggers \cite{ABL}. Hence, the poisoned network can confidently predict poisoned samples as the target label in early epochs but hesitates about the benign samples.

Based on this observation, we propose a novel secure training framework, \emph{The Victim and The Beneficiary (V\&B)}, that can directly train clean models on poisoned data without resorting to benign samples. Firstly, in warming up stage, the training dataset is recomposed into a suspicious set and a credible set according to the prediction entropy is lower or higher than the predefined threshold. Then the Victim network is trained iteratively with the suspicious set to obtain a strong poisoned sample detector. Secondly, in the clean training stage, the Beneficiary network is trained with credible samples filtered by the Victim to inhibit backdoor injection. In later training epochs, the Victim network is fully exploited to provide its learned knowledge for the Beneficiary network to help the latter erase potential backdoors through semi-supervised learning. To diminish the threat of missed poisoned samples, we design a strong data augmentation \emph{AttentionMix}, which mixes the image region with high activation values according to the attention map. Compared with existing augmentations \cite{attentivecutmix, cutmix, mixup}, it has a stronger inhibition effect against stealthy backdoor attacks \cite{Dynamic, wanet, CL}.


The main contributions of this paper are as follows: \textbf{(1)} We propose a novel dual-network secure training framework, V\&B, which can train clean models on the poisoned dataset without resorting to benign samples. 
\textbf{(2)} We design a powerful data augmentation method called AttentionMix, which has a stronger inhibition effect against stealthy backdoor attacks. \textbf{(3)} Extensive experiments on two popular benchmark datasets demonstrate the effectiveness and robustness of our framework.

\section{Related Work}
\subsection{Data Augmentation}
Data Augmentation is a common-used technique to improve the generalization capabilities of DNNs. In addition to some popular weak data augmentations (such as cropping, flipping, rotation, etc.), many strong data augmentation techniques \cite{attentivecutmix, cutmix, mixup} have been proposed to further improve model performance. Mixup \cite{mixup} linearly interpolates any two images and their labels, then trains the model with generated samples. However, the generated samples tend to be unnatural, which will decrease model performances on localization and object detection tasks. Instead of linear interpolation, Cutmix \cite{cutmix} replaces a random patch with the same region from another image and mixes their labels according to the ratio of the patch area. Further, Attentive Cutmix \cite{attentivecutmix} only pastes the influential region located by an attention mechanism to other images, effectively forcing models to learn discriminative details. Yet it also mixes labels based on the ratio of the region area, ignoring the importance of the region in two original images, and direct pasting may bring complete triggers into benign samples. Our AttentionMix takes into account the importance of the region in both images and blends the influential region with the same area in another image, which can destroy the completeness of triggers.

\subsection{Backdoor Attack and Defense}
In this paper, we only consider poisoning-based attacks toward image classification that can occur during the data preparation stage.
\flushleft{\textbf{Poisoned-based backdoor attack.}}
\justifying
In early studies, attackers usually adopt simple and visible patterns as the trigger, such as a black-white checkerboard \cite{badnets}. To escape human inspections, some works use human-imperceptible deformations \cite{wanet} as the trigger or design unique triggers \cite{li2021invisible, Dynamic} for each poisoned sample. However, they randomly select benign samples to inject triggers and change their labels (called poison-label attacks), possibly causing the content of an image not to match its label. Instead of randomly selecting samples to poison, Turner \etal~\cite{CL}, Mauro \etal~\cite{SIG}, and Liu \etal~\cite{reflection} only inject triggers into benign samples under the target label (called clean-label attacks). 
Although some triggers are imperceptible in the input space, they can be easily detected by defenses focused on latent space extracted by CNNs. Therefore, recent studies \cite{DBLP:conf/cvpr/Zhao0XDWL22, imperceptible} not only ensure the trigger's invisibility in the input space but also restrict the similarity between poison samples and corresponding benign samples in the latent space.


\flushleft{\textbf{Backdoor Defense.}}
\justifying
For poisoning-based attacks, defense methods could be divided into three categories from their working stage, including \textbf{(1)} preprocessing before training \cite{doan2020februus, hayase2021spectre, DBLP:conf/asiaccs/0001ZGZQT21}, \textbf{(2)} poison suppression during training \cite{DBD, ABL, DBLP:conf/icpr/LiuLWL20}, and \textbf{(3)} backdoor erasing after training \cite{NAD, neuralclense, DBLP:conf/iclr/ZhaoCDRL20}. For preprocessing, 
Hayase \etal~\cite{hayase2021spectre} and Chen \etal~\cite{chen2018detecting} filter the training set through statistics or clustering and train the model with benign samples. For poison suppressing, Borgnia \etal~\cite{borgnia2021strong} leverage strong data augmentation (\eg\ Mixup, Cutmix) to alleviate poisoned samples' adverse effects. Huang \etal~\cite{DBD} design a decoupling training process, which first learns a feature extractor via SimCLR \cite{SimCLR} to prevent poisoned samples from clustering in feature space. 
For backdoor erasing, Liu \etal~\cite{FP} prune the low-activation neurons for validation samples (benign samples) and then fine-tune the model with these benign samples. While Li \etal~\cite{NAD} first fine-tune the poisoned model with a small portion of benign samples to obtain a teacher model, then erase the original model's backdoor through knowledge distillation.

\subsection{Semi-supervised Learning}
Semi-supervised learning \cite{chen2022semi} aims to alleviate the dependence of deep learning on extensive labeled training data and jointly learns the model from a limited amount of labeled samples and a large number of unlabeled samples. Existing methods \cite{mixmatch, weston2008deep, zhou2003learning} generally assign each unlabeled sample with a pseudo label and design different loss functions for labeled and unlabeled samples. The weight of unlabeled sample loss will gradually increase during training to reduce the negative effect of low-quality pseudo labels in early epochs.

\section{Method}
\subsection{Preliminaries}
\flushleft{\textbf{Pipeline of Poisoned-based Backdoor Attacks.}}
\justifying
Let $\mathcal{D}_B=\{(\boldsymbol{x}_i, y_i)\}_{i=1}^N$ denotes the benign training set containing $N$ samples, where $\boldsymbol{x}_i \in \{0, 1, ..., 255\}^{C \times H \times W}$ is the image, $y_i \in \{0, 1, ..., K\}$ is its label, $K$ is the number of classes, $H$, $W$ and $C$ are the height, width and channels of an image. The attacker will select a subset of benign samples to poison with a designed injection function $G(\cdot)$. Let $\mathcal{D}_G$ indicates the generated poisoned samples set and $\mathcal{D}_R$ denotes the remaining benign samples set, where $\mathcal{D}_G = \{(\boldsymbol{x}^{\prime}, y_t) | \boldsymbol{x}^{\prime} = G(\boldsymbol{x}), (\boldsymbol{x}, y) \in \mathcal{D}_B\setminus \mathcal{D}_R \}$, $y_t$ is the target label and $r=\frac{|\mathcal{D}_G|}{\mathcal{D}_B}$ is the poisoning rate. The poisoned dataset $\mathcal{D}_P = \mathcal{D}_G \cup \mathcal{D}_R$ will be unknowingly adopted by victim users to train DNNs. At inference time, the attacker can manipulate the poisoned model's output through the injected backdoor.

\flushleft{\textbf{Threat Model and Defense Setting.}} 
\justifying
In this paper, we focus on defending against poisoned-based backdoor attacks toward image classification. The attacker can arbitrarily modify the training set, but can not change other training components (\eg\ model structure and hyperparameters) \cite{SIG, badnets, wanet}. Such attacks may occur when using third-party datasets or training platforms that can poison users' clean data. We follow the common assumption \cite{hayase2021spectre, DBD, ABL} that defenders have no information about the backdoor (including poisoning rate, trigger pattern, target label, etc.) and any local benign samples to use but have full control over the training process. The defense goal is to prevent the trained DNNs from predicting poisoned samples as the target label while maintaining its accuracy on benign samples.

\begin{figure*}[ht]
\begin{center}
\includegraphics[width=1.0\linewidth]{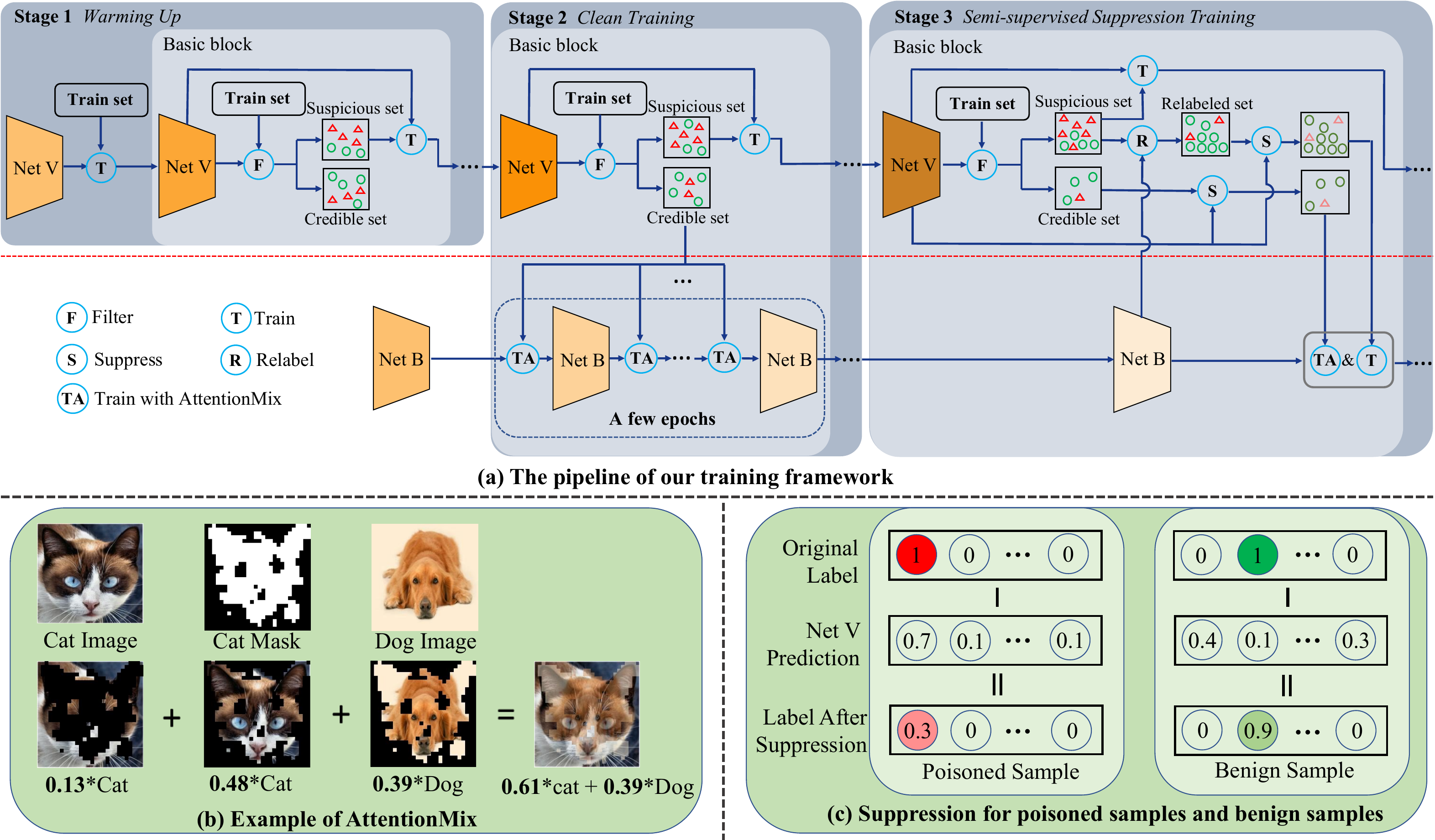}
\end{center}
   \caption{(a) The pipeline of our training framework. Net V is the Victim network (\ie\ poisoned network) that only learns from the filtered suspicious set, and Net B is the Beneficiary network (\ie\ clean network) we wanted. Red triangles represent poisoned samples, and green circles are benign samples. The color of nets indicates their degree of poisoning: the darker the color, the deeper the poisoning, and vice versa. 
   Deep blue areas are three training stages, and light blue areas are basic blocks that will be executed many times in the corresponding stage. (b) An example of AttentionMix. The bottom four images are the non-activation region of the cat image, the activation region of the cat image, the same activation region in the dog image, and the final generated image, respectively. And their labels are located below the images. (c) Suppression for poisoned samples and benign samples. The target label is assumed as the first label. Suppression can reduce the target label component of poisoned samples but have little influence on the ground-truth label of benign samples. }
   \vspace{-0.3cm}
\label{fig: overview}
\end{figure*}
 
\subsection{Overview}
In this subsection, we briefly introduce the pipeline of our framework. As Figure~\ref{fig: overview}(a) illustrates, our framework consists of two networks, and the whole training process is divided into three stages --- warming up, clean training and semi-supervised suppression training. In warming up, we aim to obtain a poisoned network that can roughly distinguish poisoned samples from benign ones according to the prediction entropy. Therefore we train the Victim network with only filtered suspicious samples (\ie\ suspicious set) except for the first epoch. This training strategy remains unchanged in the following two stages. During clean training, we adopt credible samples (\ie\ credible set) to train the Beneficiary network with our proposed AttentionMix data augmentation. The strong data augmentation can effectively prevent missed poisoned samples from creating backdoors. As the Victim network is trained, the entropy of its prediction for benign samples will also slowly decrease below the filtering threshold, leading to a reduction in the credible set size. So one credible set will be used for a few epochs to ensure the Beneficiary network gets enough training samples.

To further improve the Beneficiary network performance and erase its potential backdoor, we regard the credible set as a labeled dataset and the suspicious set as an unlabeled dataset to train the Beneficiary network via semi-supervised learning. Following conventional practice \cite{mixmatch, li2020dividemix}, each unlabeled sample will be assigned a pseudo label generated by the Beneficiary network, forming the relabeled set. However, if the Beneficiary network still contains a backdoor after stage 2, poisoned samples in the suspicious set will be relabeled as the target label, strengthening the backdoor. After specific training, the Victim network is sensitive to trigger patterns and thus can be utilized to suppress the component corresponding to the target class after relabeling. Furthermore, we also apply this suppression operation to the credible set to further weaken the impact of missed poisoned samples.

\subsection{Warming Up}
\label{section: warmup}
At this stage, we aim to obtain a poisoned network that has only learned trigger patterns. In this way, the entropy of its prediction for poisoned samples will be significantly lower than that of benign samples. Then we can use a threshold to filter out poisoned samples. 

Let $\mathcal{D}$ denotes the training set and $F_v:\{0, 1, ..., 255\}^{C \times H \times W} \rightarrow [0, 1]^K$ indicates the Victim network. We first use the whole training set to train the Victim network for one epoch with cross-entropy loss. Then the prediction entropy of a sample will be calculated with Equation \ref{entropy}, where $p$ is the prediction of the Victim network for that sample.
\begin{equation}
    e = -\sum_{k=1}^{K} p_k \log p_k  \label{entropy}
\end{equation}

Once all samples' prediction entropy is obtained, we normalize them to $[0, 1]$ through \emph{mix-max normalization}. Then a threshold $t_f$ is set to filter out suspicious samples $\mathcal{D}_s = \{(\boldsymbol{x}_i, y_i) | (\boldsymbol{x}_i, y_i)\in \mathcal{D}, e_i < t_f\}$ and the remaining samples are regarded as credible samples $\mathcal{D}_c$. To strengthen the learning of poisoned samples and delay the learning of benign samples, the Victim network will only be trained on suspicious sets in the future. The loss function is as follows:
\begin{equation}
    \mathcal{L}_v = \mathbb{E}_{(\boldsymbol{x},y) \in \mathcal{D}_s} [\mathcal{L}_{CE}(F_v(\boldsymbol{x}), y)]
\end{equation}
where $\mathcal{L}_{CE}$ denotes the cross-entropy loss.

After the first epoch, the prediction entropy of some poisoned samples may still be larger than that of most benign ones, although there is a large gap between their average values in Figure \ref{fig: entropy}. If the filtering threshold $t_f$ is set too large, lots of benign samples will be discarded, which will affect the model performance on benign samples. So we set a smaller $t_f$ and wait for the prediction entropy of these poisoned samples to decrease. As training progresses, their prediction entropy will drop below $t_f$, making the credible set cleaner. Simultaneously, the prediction entropy of benign samples will also decrease, resulting in a smaller credible set. We expect the Victim network to filter out a clean and relatively large credible set after this stage.

\subsection{Clean Training}
At this stage, we begin to train the Beneficiary network with filtered credible samples $\mathcal{D}_c $, where the filtering strategy is the same as in stage 1. Because of the rough separation, a few poisoned samples may be included in the credible set. For strong attacks, a small number of poisoned samples is enough to create a backdoor. So we design a strong data augmentation called AttentionMix to inhibit the missed poisoned samples from working. Compared with existing augmentations \cite{attentivecutmix, cutmix, mixup}, AttentionMix has a stronger inhibition effect on stealthy backdoor attacks.

The core idea of AttentionMix is to generate a new training sample $(\tilde{\boldsymbol{x}}, \tilde{\boldsymbol{y}})$ by mixing two training samples $(\boldsymbol{x}_1, y_1)$ and $(\boldsymbol{x}_2, y_2)$ according to their attention maps, where $\tilde{\boldsymbol{y}}$ is a K-dimensional vector. Following \cite{NAD}, we adopt one sample's final feature maps to compute its attention map:
\begin{equation}
    \boldsymbol{A}^{\prime} = \sum_{i=1}^{c} |F_i|^2
\end{equation}
where $\boldsymbol{F} \in \mathbb{R}^ {c \times h \times w}$ indicates feature maps, $\boldsymbol{A}^{\prime} \in \mathbb{R}^{h \times w}$ is the corresponding attention map, $c$, $h$ and $w$ are the dimensions of channel, height and width. For each attention map, we first bilinearly interpolate it to the original image size $H \times W$, and then normalize its elements to $[0, 1]$ by \emph{min-max normalization}.

After getting the attention map, we set a threshold $t_m$ to obtain the activation region that we want to mix:
\begin{equation}
\boldsymbol{M}_{i,j} =\left\{
\begin{array}{cl}
0  &  \boldsymbol{A}_{i,j} < t_m \\
1  &  \boldsymbol{A}_{i,j} \geq t_m \\
\end{array} \right.
\end{equation}
where $\boldsymbol{M} \in \mathbb{R}^{H \times W}$ is the binary mask indicating which pixels belong to the activation region and $\boldsymbol{A}$ is the attention map after interpolation and normalization.

Let $\boldsymbol{A}^{1}$ and $\boldsymbol{A}^{2}$ denote the attention maps of $\boldsymbol{x}_1$ and $\boldsymbol{x}_2$, $\boldsymbol{M}^{1}$ denotes the first sample's mask. As Figure \ref{fig: overview}(b) shows, the mixed image $\tilde{\boldsymbol{x}}$ is composed of three parts: the non-activation region of $\boldsymbol{x}_1$, the activation region of $\boldsymbol{x}_1$ and the same area in $\boldsymbol{x}_2$. During mixing, the non-activation region of $\boldsymbol{x}_1$ remains unchanged, and the activation region is blended with the same area of $\boldsymbol{x}_2$:
\begin{equation}
\begin{split}
\tilde{\boldsymbol{x}} = (\textbf{1} - \boldsymbol{M}^1) \odot \boldsymbol{x}_1 + \lambda_2 \cdot (\boldsymbol{M}^1 \odot \boldsymbol{x}_1) \\ + \lambda_3 \cdot (\boldsymbol{M}^1 \odot \boldsymbol{x}_2) \label{Eq 5}
\end{split}
\end{equation}
where $\odot$ is the Hadamard Product, $\lambda_2$ and $\lambda_3$ are the mixing weights calculated from the two attention maps. The calculation process is as follows:
\begin{equation}
w_1 = \frac{{\rm sum}(\boldsymbol{M}^{1} \odot \boldsymbol{A}^{1} )}{{\rm sum}(\boldsymbol{A}^{1})}, 
w_2 = \frac{{\rm sum}(\boldsymbol{M}^{1} \odot \boldsymbol{A}^{2} )}{{\rm sum}(\boldsymbol{A}^{2})} \label{Eq 6}
\end{equation}
\begin{equation}
\lambda_1 = 1-w_1, \lambda_2 = \frac{w_1}{w_1+w_{2}}, \lambda_3 = \frac{w_{2}}{w_1+w_{2}}    \label{Eq 7}
\end{equation}
where ${\rm sum(\cdot)}$ is a summation function, $w_1$ and $w_2$ denote the importance of the activation region to the two images, and $\lambda_1$ is the importance of the non-activation region to $\boldsymbol{x}_1$. The final label is also composed of three parts:
\begin{equation}
    \tilde{\boldsymbol{y}} = \lambda_1 \cdot \boldsymbol{y}_1 + w_1 \cdot (\lambda_2 \cdot \boldsymbol{y}_1 + \lambda_3 \cdot \boldsymbol{y}_2)  \label{Eq 8}
\end{equation}
where $\boldsymbol{y}_1$ and $\boldsymbol{y}_2$ are the one-hot labels of the two samples. When the activation region of $\boldsymbol{x}_1$ corresponds to the background of $\boldsymbol{x}_2$, $w_1$ will be significantly larger than $w_2$, and $\lambda_3$ will be a small value, which can reduce background interference during mixing.

We denote the generated new credible set as $\tilde{\mathcal{D}}_c$ and train the Beneficiary network $F_b$ with the following loss function (\ie\ \emph{Train with AttentionMix} in Fiqure \ref{fig: overview}(a)):
\begin{equation}
    \mathcal{L}_b = \mathbb{E}_{(\tilde{\boldsymbol{x}}, \tilde{\boldsymbol{y}}) \in \tilde{\mathcal{D}}_c} [\mathcal{L}_{CE}(F_b(\tilde{\boldsymbol{x}}), \tilde{\boldsymbol{y}})]
\end{equation}

For a poisoned network, the activation region of poisoned samples is more likely their trigger locations. Mixing the activation region can destroy the completeness of trigger patterns. While for benign samples, the mixing plays a regularization role that prevents the network from overfitting a particular subject and forces it to learn abundant features.

\begin{table*}[htbp]
\begin{center}
\LARGE
\resizebox{\linewidth}{!}{
    \begin{tabular}{c|cccccccc|cccccc}
    \toprule[2pt]
    Attack & \multicolumn{2}{c}{BadNets \cite{badnets}} & \multicolumn{2}{c}{Blend \cite{blend}} & \multicolumn{2}{c}{Dynamic \cite{Dynamic}} & \multicolumn{2}{c|}{WaNet \cite{wanet}} & \multicolumn{2}{c}{SIG \cite{SIG}} & \multicolumn{2}{c}{CL-16 \cite{CL}} & \multicolumn{2}{c}{\text {CL-32 \cite{CL}}} \\
    \hline
    Metric & BA    & ASR   & BA    & ASR   & BA    & ASR   & BA    & ASR   & BA    & ASR   & BA    & ASR   & BA    & ASR \\
    \hline
    No Defense & 94.36\% & 100.00\% & 94.57\% & 99.76\% & 94.66\% & 95.18\% & 94.21\% & 95.84\% & 95.09\% & 64.79\% & 94.98\% & 96.93\% & 95.03\% & 94.73\% \\
    \hline
    FP \cite{FP}    & \underline{92.72\%} & 1.92\% & \underline{92.49}\% & 10.61\% & \underline{92.47\%} & 11.73\% & \underline{93.27\%} & 0.90\% & \underline{93.55\%} & 37.98\% & \underline{92.92\%} & 9.74\% & 93.03\% & 6.39\% \\
    NAD \cite{NAD}  & 88.83\% & 1.83\% & 87.72\% & \underline{1.38\%} & 88.30\% & \underline{2.32\%} & 91.00\% & 0.99\% & 90.42\% & 3.52\% & 88.92\% & 1.64\% & 90.19\% & \underline{2.60\%} \\
    \hline
    ABL \cite{ABL}  & 90.47\% & \underline{0.70\%} & 91.31\% & 31.49\% & 83.07\% & 94.21\% & 80.78\% & 19.93\% & 88.61\% & \underline{1.37}\% & 90.29\% & \textbf{0.54\%} & 86.50\% & 2.68\% \\
    DBD \cite{DBD}  & 92.00\% & 3.06\% & 91.62\% & 3.78\% & 90.98\% & 17.56\% & 89.13\% & \textbf{0.11\%} & 89.42\% & 1.61\% & 92.85\% & 91.08\% & \underline{93.24\%} & 14.88\% \\
    V\&B (Ours) & \textbf{93.96\%} & \textbf{0.62\%} & \textbf{94.37\%} & \textbf{0.63\%} & \textbf{93.91\%} & \textbf{1.13\%} & \textbf{94.15\%} & \underline{0.54\%} & \textbf{94.08\%} & \textbf{0.17\%} & \textbf{94.24\%} & \underline{1.01\%} & \textbf{93.98\%} & \textbf{0.64\%} \\
    \bottomrule[2pt]
    \end{tabular}
}
\end{center}
\caption{The performance of 5 defense methods against 7 backdoor attacks on CIFAR-10. The first 4 attacks are poison-label attacks and the last 3 attacks are clean-label attacks. Note that FP and NAD require local benign samples to work, while ABL, DBD and our method do not. The best results are marked in boldface and underlined items are the second-best results.}
\label{Table1}
\end{table*}

\begin{table*}[htbp]
\Huge
\begin{center}
\resizebox{0.6\linewidth}{!}{
    \begin{tabular}{c|cccccc|cc}
    \toprule[3pt]
    Attack & \multicolumn{2}{c}{BadNets \cite{badnets}} & \multicolumn{2}{c}{Blend \cite{blend}} & \multicolumn{2}{c|}{WaNet \cite{wanet}} & \multicolumn{2}{c}{SIG \cite{SIG}} \\
    \hline
    Metric & BA    & ASR   & BA    & ASR   & BA    & ASR   & BA    & ASR \\
    \hline
    No Defense & 94.94\% & 99.51\% & 95.16\% & 100.00\% & 92.63\% & 100.00\% & 90.38\% & 89.22\% \\
    \hline
    FP \cite{FP}   & 86.28\% & 1.05\% & 87.63\% & 2.03\% & 82.82\% & 26.19\% & 82.50\% & 22.76\% \\
    NAD \cite{NAD}  & 85.83\% & 3.82\% & 86.73\% & 2.90\% & 83.93\% & 3.88\% & 80.71\% & 3.88\% \\
    \hline
    ABL \cite{ABL}  & 92.15\% & 13.78\% & 87.34\% & 56.57\% & 91.38\% & 56.71\% & 80.13\% & \underline{0.87\%} \\
    DBD \cite{DBD}  & \underline{93.08\%} & \underline{0.31}\% & \underline{93.81\%} & \underline{0.87\%} & \underline{92.06\%} & \textbf{1.00\%} & \underline{84.47\%} & 98.47\% \\
    V\&B (Ours) & \textbf{95.42\%} & \textbf{0.28\%} & \textbf{95.03\%} & \textbf{0.45\%} & \textbf{94.84\%} & \underline{1.92\%}   & \textbf{94.65\%} & \textbf{0.03\%} \\
    \bottomrule[3pt]
    \end{tabular}
}
\end{center}
\caption{The performance of 5 defense methods against 4 backdoor attacks on ImageNet subset.}
\vspace{-0.4cm}
\label{Table2}
\end{table*}


\subsection{Semi-supervised Suppression Training}
Until now, we have got a poisoned Victim network that can filter out most of the poisoned samples and a Beneficiary network that is relatively clean but may still contain the backdoor. In the previous stage, the Beneficiary network is trained only with partial training samples (\ie\ the credible set), which may harm its performance on benign samples. Thus we intend to enable the deprecated suspicious samples via semi-supervised learning at this stage to improve the Beneficiary network while erasing its potential backdoor.

Given a credible set $\mathcal{D}_c$ and a suspicious set $\mathcal{D}_s$, we use samples in $\mathcal{D}_c$ as labeled samples and remove the labels of samples in $\mathcal{D}_s$ to transform the problem into a semi-supervised learning setting. For each unlabeled sample, we average the predictions of the Beneficiary network over its multiple augmentations (\eg\ random crop and random flip) and select the class with max probability as its pseudo-label, which is similar to \cite{mixmatch} and \cite{li2020dividemix}. The relabeled samples are denoted as $\mathcal{D}_r = \{(\boldsymbol{x}_i, \hat{y}_i) | (\boldsymbol{x}_i, y_i) \in \mathcal{D}_s \}$.

However, the poisoned samples in $\mathcal{D}_s$ may be relabeled as the target label, \ie\ $\hat{y}_i=y_t$, if the Beneficiary network still contains the backdoor. We can mitigate the impact of such mislabeling by exploiting the knowledge learned by the Victim network. Specifically, we adopt the Victim network's prediction to suppress the pseudo-label (\ie\ \emph{Suppress} in Figure \ref{fig: overview}(a)):
\begin{equation}
    \boldsymbol{y}_{i}^{\prime} = {\rm clip}(\hat{\boldsymbol{y}_i} - {\rm softmax}(F_v(\boldsymbol{x}_i)) , 0, 1)
\end{equation}
where $(\boldsymbol{x}_i, \hat{y}_i) \in \mathcal{D}_r$, 
$\hat{\boldsymbol{y}}_i$ is the one-hot vector of $\hat{y}_i$ and ${\rm clip} (\cdot,0,1)$ is a function that will restrict its inputs to $[0, 1]$ (\ie\ inputs smaller than 0 became 0, inputs larger than 1 became 1, and the rest remain unchanged).

We use $\mathcal{D}_r^{\prime} = \{(\boldsymbol{x}_i, \boldsymbol{y}_i^{\prime}) | (\boldsymbol{x}_i, \hat{y}_i) \in \mathcal{D}_r \}$ to indicate the suspicious samples after relabeling and suppression. We also suppress the labels of credible samples in the same way to weaken the influence of poisoned samples that may be missed after filtering. As in stage 2, the suppressed credible samples are augmented with our AttentionMix, forming the training set $\tilde{\mathcal{D}}_c^{\prime}$ to further inhibit backdoor injection. We fine-tune the Beneficiary network with the following loss function:
\begin{equation}
\begin{split}
    \mathcal{L} = \mathbb{E}_{(\tilde{\boldsymbol{x}}, \tilde{\boldsymbol{y}}^{\prime}) \in \tilde{\mathcal{D}}_c^{\prime}} [\mathcal{L}_{CE}(F_b(\tilde{\boldsymbol{x}}), \tilde{\boldsymbol{y}}^{\prime})] + \\
    \alpha \cdot \mathbb{E}_{(\boldsymbol{x}, \boldsymbol{y}^{\prime}) \in \mathcal{D}_r^{\prime}} [\mathcal{L}_{CE}(F_b(\boldsymbol{x}), \boldsymbol{y}^{\prime})]
\end{split}
\end{equation}

Since the Victim network is only trained with suspicious samples, it will confidently predict poisoned samples as the target label but give a wrong prediction for benign samples with high probability. Thus for a poisoned sample, the suppression can reduce the component corresponding to the target class in its label while having little influence on the component of a sample's ground-truth class, as shown in Figure \ref{fig: overview}(c).

\section{Experiments}
\subsection{Experimental Settings}
\flushleft{\textbf{Datasets and Models.}}
\justifying
We employ ResNet-18 \cite{ResNet} as our Victim and Beneficiary networks, a standard setting in previous studies \cite{DBD}. For a fair comparison, we evaluate all defenses on two popular benchmark datasets, including CIFAR-10 and a subset of ImageNet containing 12 classes generated by Li \etal~\cite{ABL}. 
The details are summarized in Appendix A.1.
\begin{table*}[htbp]
\Huge
\begin{center}
\resizebox{0.73\linewidth}{!}{
    \begin{tabular}{c|cc|cc|cc|cc|cc}
    \toprule[3pt]
    Attack & \multicolumn{2}{c|}{BadNets \cite{badnets}} & \multicolumn{2}{c|}{Blend \cite{blend}} & \multicolumn{2}{c|}{Dynamic \cite{Dynamic}} & \multicolumn{2}{c|}{WaNet \cite{wanet}} & \multicolumn{2}{c}{CL-16 \cite{CL}} \\
    \hline
    Metric & BA    & ASR   & BA    & ASR   & BA    & ASR  & BA    & ASR   & BA    & ASR \\
    \hline
    No Defense & 94.36\% & 100\% & 94.57\% & 99.76\% & 94.66\% & 95.18\% & 94.21\% & 95.84\% & 94.98\% & 96.93\% \\
    \hline
    Mixup \cite{mixup} & 93.37\% & \textbf{0.27\%} & 92.80\% & 3.49\% & 92.97\%  & 42.10\%  & 92.97\% & 42.10\% & \underline{94.80\%} & 20.03\% \\
    Cutmix \cite{cutmix}& 93.52\% & \underline{0.40\%} & 93.40\% & \underline{1.07\%} & 91.40\% & 53.23\% & 91.40\% & 53.23\% & 94.54\% & \underline{1.16\%} \\
    Attentive Cutmix \cite{attentivecutmix}& \textbf{94.26\%} & \underline{0.40\%} & \textbf{94.55\%} & 4.62\% &  \underline{93.09\%} & \underline{13.38\%} & \underline{93.09\%} & \underline{13.88\%} & \textbf{95.18\%} & 2.97\% \\
    AttentionMix (Ours) & \underline{93.96\%} & 0.62\% & \underline{94.37\%} & \textbf{0.63\%} & \textbf{93.91\%} & \textbf{1.13\%} & \textbf{94.15\%} & \textbf{0.54\%} & 94.24\% & \textbf{1.01\%} \\
    \bottomrule[3pt]
    \end{tabular}%
}
\end{center}
\caption{The inhibition effect of 4 strong data augmentations against 5 backdoor attacks on CIFAR-10. 
}
\vspace{-0.4cm}
\label{Table3}
\end{table*}
\flushleft{\textbf{Attack Configures.}}
\justifying
We consider 6 representative backdoor attacks in our experiments, including four poison-label attacks: BadNets \cite{badnets}, Blend attack \cite{blend}, WaNet attack \cite{wanet}, Dynamic attack \cite{Dynamic}, and two clean-label attacks: Sinusoidal signal attack (SIG) \cite{SIG} and Clean-label attack (CL) \cite{CL}. 
To achieve the best attack performance, we configure these attacks as close as possible to their released codes and original settings (\eg\ trigger pattern and trigger size). While on the ImageNet subset, we implement these attacks based on the settings suggested by \cite{DBD}, since some attacks miss the corresponding experiments in their original papers. We uniformly set the poisoning ratio as 0.1 and select the first class as the target class in both datasets, \ie\ $y_t = 0$. More details about attacks are shown in Appendix A.2.

\flushleft{\textbf{Defense Configures and Training Details.}}
\justifying
We compare our V\&B with 4 state-of-the-art defenses: Fine-pruning (FP) \cite{FP}, Neural Attention Distillation (NAD) \cite{NAD}, Anti-Backdoor Learning (ABL) \cite{ABL} and Decoupling (DBD) \cite{DBD}, in which FP and NAD require additional local benign samples to work. Similarly, we follow their original settings and released codes. For V\&B, we warm up the Victim network for 3 epochs on CIFAR-10 and 6 epochs on the ImageNet subset, where the filtering threshold $t_f$ linearly decreases from 1 to 0.2 and from 1 to 0.4. The selection of these hyperparameters and an automatic warming-up strategy can be found in Appendix A.3. In the following stages, the filtering threshold keeps unchanged (\ie\ 0.2 for CIFAR-10 and 0.4 for ImageNet subset). 
For fast convergence, we adopt the Adam optimizer to train the Victim network with a learning rate of 0.001 in the whole training process. We optimize the Beneficiary network for 200 epochs using an SGD optimizer with a momentum of 0.9 and a weight decay factor of $10^{-4}$, where the first 100 epochs belong to stage 2 and the last 100 epochs belong to stage 3. The learning rate is initially set to 0.1 and is divided by 10 after every 50 epochs. In particular, the Victim network is optimized every 5 epochs at stage 2, \ie\ one credible set is used to train the Beneficiary network for 5 epochs. We use a batch size of 128 for CIFAR-10 and 16 for the ImageNet subset, and set the mixing threshold $t_m$ to 0.1 in all cases. Following semi-supervised learning, we gradually increase the weight of suspicious sample loss in stage 3, simply setting $\alpha$ to 0.1 in the first 50 epochs and 0.5 in the last 50 epochs.

\flushleft{\textbf{Evaluation Metrics.}}
\justifying
We evaluate the performance of defenses with two commonly used metrics: the model's classification accuracy on the benign test set, Benign Accuracy (BA), and the model's classification accuracy on the poison test set, Attack Success Rate (ASR). The lower the ASR and the higher BA, the stronger the defense. In particular, the poison test set removes the samples whose ground-truth label is the target label.

\subsection{Effectiveness of Our V\&B Defense}
To measure the effectiveness of our proposed V\&B method, we compare its performance with four state-of-the-art defenses on CIFAR-10 and the ImageNet subset. The backdoor attack CL-16 and CL-32 are two versions of the Clean-label attack with different magnitudes of adversarial perturbations.

Table \ref{Table1} demonstrates the defense results on CIFAR-10, where \emph{No Defense} is a baseline representing the capability of backdoor attacks. Considering real scenarios, we adopt common data augmentations (\ie\ random crop and horizontal flipping) during training, which will slightly affect the attack success rate. It is obvious that our V\&B achieves the best results against almost all attacks, even compared to defenses that need additional benign samples. Our method reduces the success rate of all attacks from above $95\%$ ($64.79\%$ for SIG) to blow or close to $1\%$, while maintaining the accuracy on benign samples with only about $1\%$ reduction. For WaNet and CL-16, V\&B achieves the second-best performance in reducing the attack success rate, but the benign accuracy of V\&B is significantly superior to the corresponding best defense, \ie\ DBD and ABL. WaNet adopts a warping field as the trigger and adds a noisy training mode to interfere with detection. CL-16 only poisons the samples whose ground-truth label is the target label and adds adversarial perturbations to them. So it is relatively hard to distinguish the poisoned samples they craft from the benign ones.

The defense performances on the ImageNet subset are shown in Table \ref{Table2}. Compared with other defenses, our method has the best accuracy on benign samples against all four attacks, even outperforming the baseline due to our AttentionMix data augmentation and semi-supervised suppression training. Semi-supervised learning can relabel poisoned samples to their ground-truth labels, improving benign accuracy. In terms of reducing the attack success rate, our method achieves the best results against BadNets, Blend, and SIG, and the second-best result against WaNet, which is comparable to the best result ($1.92\%$ vs. $ 1.00\%$).

We also validate the effectiveness of our method under different poisoning rates in Appendix A.4. Even with a poisoning rate of 50\%, our method can still reduce the attack success rate of most attacks to below 1\%, while maintaining a satisfactory benign accuracy.

\begin{table*}[htbp]
\large
\begin{center}
\resizebox{1\linewidth}{!}{
    \begin{tabular}{c|cccccccc|cc}
    \toprule[2pt]
    Attack & \multicolumn{2}{c}{BadNets \cite{badnets}} & \multicolumn{2}{c}{Blend \cite{blend}} & \multicolumn{2}{c}{WaNet \cite{wanet}} & \multicolumn{2}{c|}{Dynamic \cite{Dynamic}} & \multicolumn{2}{c}{CL-16 \cite{CL}} \\
    \hline
    Metric & BA   & ASR   & BA   & ASR   & BA   & ASR   & BA   & ASR   & BA   & ASR \\
    \hline
    No Defense & 94.36\% & 100.00\% & 94.57\% & 99.76\% & 94.21\% & 95.84\% & 94.66\% & 95.18\% & 94.98\% & 96.93\% \\
    \hline
    + F    & 93.50\% & 10.01\% & 93.47\% & 76.70\% & 92.92\% & 90.97\% & 93.80\% & 93.04\% & 93.63\% & 2.70\% \\
    + F + AT & 92.27\% & \textbf{0.42\%} & 92.72\% & 1.42\% & 91.58\% & 25.16\% & 93.08\% & 50.01\% & 93.09\% & 2.09\% \\
    + F + AT + SST & 93.88\% & 1.21\% & \textbf{94.56\%} & 1.52\% & 93.31\% & 1.23\% & \underline{94.32\%} & 28.79\% & \textbf{94.43\%} & 3.29\% \\
    + F + AT + SST + S (suspicious) & \textbf{94.11\%} & 1.00\% & \underline{94.49\%} & \underline{0.73\%} & \underline{93.49\%} & \underline{0.61\%} & \textbf{94.61\%} & \underline{1.43\%} & 94.00\% & \underline{1.06\%} \\
    ALL & \underline{93.96\%} & \underline{0.62\%} & 94.37\% & \textbf{0.63\%} & \textbf{94.15\%} & \textbf{0.54\%} & 93.91\% & \underline{\textbf{1.13\%}} & \underline{94.24}\% & \textbf{1.01\%} \\
    \bottomrule[2pt]
    \end{tabular}%
}
\end{center}
\caption{Ablation study of individual components in our framework on CIFAR-10. }
\label{Table4}
\end{table*}

\subsection{Effectiveness of AttentionMix}
To illustrate the inhibition effect of our AttentionMix, we replace it with three other popular strong data augmentations in our framework, \ie\ Mixup, Cutmix and Attentive Cutmix. As Table \ref{Table3} shows, each data augmentation performs well against BadNets and Blend, while for more stealthy attacks, such as WaNet and Dynamic, our AttentionMix can better reduce the attack success rate. This is because most of poisoned samples crafted by BadNets and Blend can be filtered out before training the Beneficiary network due to the simple trigger patterns. But for stealthy attacks, the Victim network will only separate part of the poisoned samples after warming up, and the other three augmentations can not effectively inhibit the missed poisoned samples. For a poisoned sample, the poisoned network will focus on its trigger location, so blending the sample's activation region with other images can effectively destroy triggers and help to clean the network. Although Attentive Cutmix can locate the trigger region, direct pasting may leave the trigger intact in the generated image. 

\subsection{Ablation Studies}
\flushleft{\textbf{Component Effects.}}
\justifying
As shown in Table \ref{Table4}, We first train the network without any defense as a baseline and then add our components one by one to investigate their effect. The variants are as follows: (1) + F: filtering the training set and training on credible samples; (2) + F + AT: using AttentionMix when training with credible samples; (3) + F + AT + SST: adding standard semi-supervised training \cite{mixmatch} on the basis of the second variant; (4) + F + AT + SST + S (suspicious): suppressing suspicious samples during semi-supervised training on the basis of the third variant.  

As can be seen from Table \ref{Table4}, when defending against Blend, WaNet, and Dynamic, only filtering will fail because enough poisoned samples are missed to create backdoors. After using AttentionMix, although the attack success rate of most attacks is severely decreased, the benign accuracy also drops slightly. Thus semi-supervised training is adopted to improve the performance of the network on benign samples. For WaNet and Dynamic, semi-supervised training also reduces their attack success rate by relabeling poisoned samples as ground-truth labels. However, the Beneficiary network is not completely clean, so a small number of poisoned samples may be relabeled as the target label, causing the attack success rate to rebound (\eg\ BadNets, Blend and Cl-16). Suppression to suspicious samples can make up for the weakness of relabeling and adding suppression to credible samples will further reduce the attack success rate. We say that the complete framework is more secure when facing backdoor attacks, although its benign accuracy may be lower than other variants.

To further demonstrate the effect of stage 3, we compare the final results with those after stage 2 in Appendix A.5. In all cases, the attack success rate after stage 2 is higher than the final result, while the benign accuracy after stage 2 is lower than the final result, suggesting that our semi-supervised suppression training can both erasing existing backdoors and improve model performance.

\begin{figure}[t]
\begin{center}
\includegraphics[width=1.0\linewidth]{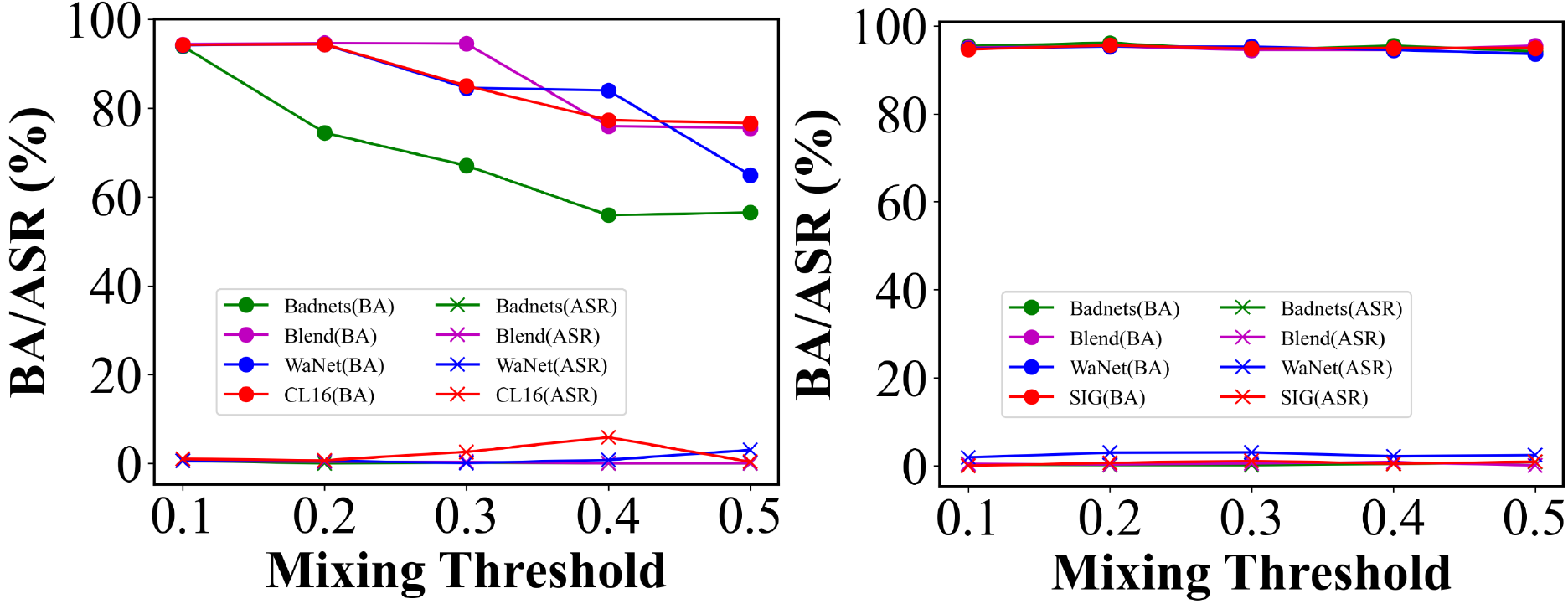}
\end{center}
   \caption{Ablation study of mixing threshold $t_m$ on CIFAR-10 (left) and ImageNet subset (right).}
\label{fig: 5}
\end{figure}

\flushleft{\textbf{Selection of Mixing Threshold.}}
\justifying
As Figure \ref{fig: 5} shows, we set the mixing threshold $t_m$ from $0.1$ to $0.5$ on CIFAR-10 and ImageNet subset to study its impact on model performance. 
It can be seen that the benign accuracy on CIFAR-10 starts to drop when the mixing threshold is large enough (\eg $0.2$ for BadNets and $0.3$ for Blend), while the attack success rate only slightly fluctuates. The decline in benign accuracy is mainly caused by the mislabeled samples in stage 3. 
Due to the small image size (32 $\times$ 32 in CIFAR-10), the activation region will contain only a few clustered pixels or some separate pixels if the mixing threshold is too large. During mixing, these pixels will carry their labels but can not provide the feature of corresponding classes, which is equivalent to introducing noise to labels. It hinders the learning of our Beneficiary network, leading to more and more suspicious samples being mislabeled in stage 3. This point can be confirmed on the ImageNet subset (image size is 224 $\times$ 224), where the mixing threshold has little impact on our method. 

\section{Conclusion}
In this paper, we show that poisoned samples could be easily distinguished by the entropy of the poisoned network's prediction. Based on this finding and the fact that strong data augmentation can inhibit backdoor injection, we propose a novel secure training framework consisting of two parallel networks to prevent a backdoor generation from three aspects: (1) filtering training data; (2) inhibiting backdoor injection; (3) erasing potential backdoors. Specifically, we sacrifice one network to exclusively learn backdoor triggers and filter credible samples according to the prediction entropy to train another network. During training, we design a strong data augmentation to inhibit missed poisoned samples from working and then introduced a semi-supervised learning method to erase potential backdoors with the help of the sacrifice network. Extensive experiments verify that our framework is effective against various backdoor attacks under different poisoning rates.

\flushleft{\textbf{Acknowledgment.}}
\justifying
This work is supported in part by the National Natural Science Foundation of China Under Grants No.62176253 and No.U20B2066.

\clearpage
{\small
\bibliographystyle{ieee_fullname}
\bibliography{egbib}

\begin{thebibliography}{10}\itemsep=-1pt

\bibitem{SIG}
Mauro Barni, Kassem Kallas, and Benedetta Tondi.
\newblock {A New Backdoor Attack in CNNS by Training Set Corruption Without Label Poisoning}.
\newblock In {\em 2019 IEEE International Conference on Image Processing}, pages 101--105. IEEE, 2019.

\bibitem{mixmatch}
David Berthelot, Nicholas Carlini, Ian Goodfellow, Nicolas Papernot, Avital Oliver, and Colin~A Raffel.
\newblock {MixMatch: A Holistic Approach to Semi-Supervised Learning}.
\newblock {\em Advances in neural information processing systems}, 32, 2019.

\bibitem{borgnia2021strong}
Eitan Borgnia, Valeriia Cherepanova, Liam Fowl, Amin Ghiasi, Jonas Geiping, Micah Goldblum, Tom Goldstein, and Arjun Gupta.
\newblock {Strong Data Augmentation Sanitizes Poisoning and Backdoor Attacks Without an Accuracy Tradeoff}.
\newblock In {\em IEEE International Conference on Acoustics, Speech and Signal Processing}, pages 3855--3859. IEEE, 2021.

\bibitem{chen2018detecting}
Bryant Chen, Wilka Carvalho, Nathalie Baracaldo, Heiko Ludwig, Benjamin Edwards, Taesung Lee, Ian Molloy, and Biplav Srivastava.
\newblock {Detecting Backdoor Attacks on Deep Neural Networks by Activation Clustering}.
\newblock {\em arXiv preprint arXiv:1811.03728}, 2018.

\bibitem{proflip}
Huili Chen, Cheng Fu, Jishen Zhao, and Farinaz Koushanfar.
\newblock {ProFlip: Targeted Trojan Attack With Progressive Bit Flips}.
\newblock In {\em Proceedings of the IEEE/CVF International Conference on Computer Vision}, pages 7718--7727, 2021.

\bibitem{SimCLR}
Ting Chen, Simon Kornblith, Mohammad Norouzi, and Geoffrey Hinton.
\newblock {A Simple Framework for Contrastive Learning of Visual Representations}.
\newblock In {\em International conference on machine learning}, pages 1597--1607. PMLR, 2020.

\bibitem{blend}
Xinyun Chen, Chang Liu, Bo Li, Kimberly Lu, and Dawn Song.
\newblock {Targeted Backdoor Attacks on Deep Learning Systems Using Data Poisoning}.
\newblock {\em arXiv preprint arXiv:1712.05526}, 2017.

\bibitem{chen2022semi}
Yanbei Chen, Massimiliano Mancini, Xiatian Zhu, and Zeynep Akata.
\newblock {Semi-Supervised and Unsupervised Deep Visual Learning: A Survey}.
\newblock {\em IEEE Transactions on Pattern Analysis and Machine Intelligence}, 2022.

\bibitem{doan2020februus}
Bao~Gia Doan, Ehsan Abbasnejad, and Damith~C Ranasinghe.
\newblock {Februus: Input Purification Defense Against Trojan Attacks on Deep Neural Network Systems}.
\newblock In {\em Annual Computer Security Applications Conference}, pages 897--912, 2020.

\bibitem{badnets}
Tianyu Gu, Brendan Dolan-Gavitt, and Siddharth Garg.
\newblock {Badnets: Identifying Vulnerabilities in the Machine Learning Model Supply Chain}.
\newblock {\em arXiv preprint arXiv:1708.06733}, 2017.

\bibitem{trojannet}
Chuan Guo, Ruihan Wu, and Kilian~Q Weinberger.
\newblock {Trojannet: Embedding Hidden Trojan Horse Models in Neural Networks}.
\newblock {\em arXiv preprint arXiv:2002.10078}, 2, 2020.

\bibitem{hayase2021spectre}
Jonathan Hayase, Weihao Kong, Raghav Somani, and Sewoong Oh.
\newblock {SPECTRE: Defending Against Backdoor Attacks Using Robust Statistics}.
\newblock {\em arXiv preprint arXiv:2104.11315}, 2021.

\bibitem{ResNet}
Kaiming He, Xiangyu Zhang, Shaoqing Ren, and Jian Sun.
\newblock {Deep Residual Learning for Image Recognition}.
\newblock In {\em Proceedings of the IEEE/CVF Conference on Computer Vision and Pattern Recognition}, pages 770--778, 2016.

\bibitem{DBD}
Kunzhe Huang, Yiming Li, Baoyuan Wu, Zhan Qin, and Kui Ren.
\newblock {Backdoor Defense via Decoupling the Training Process}.
\newblock In {\em The Tenth International Conference on Learning Representations}, 2022.

\bibitem{li2020dividemix}
Junnan Li, Richard Socher, and Steven~CH Hoi.
\newblock {DIVIDEMIX: LEARNING WITH NOISY LABELS AS SEMI-SUPERVISED LEARNING}.
\newblock {\em arXiv preprint arXiv:2002.07394}, 2020.

\bibitem{li2021invisible}
Yuezun Li, Yiming Li, Baoyuan Wu, Longkang Li, Ran He, and Siwei Lyu.
\newblock {Invisible Backdoor Attack With Sample-Specific Triggers}.
\newblock In {\em Proceedings of the IEEE/CVF International Conference on Computer Vision}, pages 16463--16472, 2021.

\bibitem{ABL}
Yige Li, Xixiang Lyu, Nodens Koren, Lingjuan Lyu, Bo Li, and Xingjun Ma.
\newblock {Anti-Backdoor Learning: Training Clean Models on Poisoned Data}.
\newblock {\em Advances in Neural Information Processing Systems}, 34:14900--14912, 2021.

\bibitem{NAD}
Yige Li, Xixiang Lyu, Nodens Koren, Lingjuan Lyu, Bo Li, and Xingjun Ma.
\newblock {Neural Attention Distillation: Erasing Backdoor Triggers from Deep Neural Networks}.
\newblock In {\em International Conference on Learning Representations}, 2021.

\bibitem{FP}
Kang Liu, Brendan Dolan-Gavitt, and Siddharth Garg.
\newblock {Fine-Pruning: Defending Against Backdooring Attacks on Deep Neural Networks}.
\newblock In {\em International Symposium on Research in Attacks, Intrusions, and Defenses}, pages 273--294. Springer, 2018.

\bibitem{DBLP:conf/icpr/LiuLWL20}
Xuankai Liu, Fengting Li, Bihan Wen, and Qi Li.
\newblock {Removing Backdoor-Based Watermarks in Neural Networks with Limited Data}.
\newblock In {\em International Conference on Pattern Recognition}, pages 10149--10156. IEEE, 2021.

\bibitem{reflection}
Yunfei Liu, Xingjun Ma, James Bailey, and Feng Lu.
\newblock {Reflection Backdoor: A Natural Backdoor Attack on Deep Neural Networks}.
\newblock In {\em European Conference on Computer Vision}, pages 182--199. Springer, 2020.

\bibitem{wanet}
Anh Nguyen and Anh Tran.
\newblock {WaNet--Imperceptible Warping-based Backdoor Attack}.
\newblock {\em arXiv preprint arXiv:2102.10369}, 2021.

\bibitem{Dynamic}
Tuan~Anh Nguyen and Anh Tran.
\newblock {Input-Aware Dynamic Backdoor Attack}.
\newblock {\em Advances in Neural Information Processing Systems}, 33:3454--3464, 2020.

\bibitem{qi2022towards}
Xiangyu Qi, Tinghao Xie, Ruizhe Pan, Jifeng Zhu, Yong Yang, and Kai Bu.
\newblock {Towards Practical Deployment-Stage Backdoor Attack on Deep Neural Networks}.
\newblock In {\em Proceedings of the IEEE/CVF Conference on Computer Vision and Pattern Recognition}, pages 13347--13357, 2022.

\bibitem{DBLP:conf/asiaccs/0001ZGZQT21}
Han Qiu, Yi Zeng, Shangwei Guo, Tianwei Zhang, Meikang Qiu, and Bhavani Thuraisingham.
\newblock {DeepSweep: An Evaluation Framework for Mitigating DNN Backdoor Attacks using Data Augmentation}.
\newblock In {\em Proceedings of the 2021 ACM Asia Conference on Computer and Communications Security}, pages 363--377, 2021.

\bibitem{CL}
Alexander Turner, Dimitris Tsipras, and Aleksander Madry.
\newblock {Label-Consistent Backdoor Attacks}.
\newblock {\em arXiv preprint arXiv:1912.02771}, 2019.

\bibitem{attentivecutmix}
Devesh Walawalkar, Zhiqiang Shen, Zechun Liu, and Marios Savvides.
\newblock {Attentive CutMix: An Enhanced Data Augmentation Approach for Deep Learning Based Image Classification}.
\newblock {\em arXiv preprint arXiv:2003.13048}, 2020.

\bibitem{neuralclense}
Bolun Wang, Yuanshun Yao, Shawn Shan, Huiying Li, Bimal Viswanath, Haitao Zheng, and Ben~Y Zhao.
\newblock {Neural Cleanse: Identifying and Mitigating Backdoor Attacks in Neural Networks}.
\newblock In {\em 2019 IEEE Symposium on Security and Privacy}, pages 707--723. IEEE, 2019.

\bibitem{weston2008deep}
Jason Weston, Fr{\'e}d{\'e}ric Ratle, and Ronan Collobert.
\newblock {Deep Learning via Semi-Supervised Embedding}.
\newblock In {\em Proceedings of the 25th international conference on Machine learning}, pages 1168--1175, 2008.

\bibitem{cutmix}
Sangdoo Yun, Dongyoon Han, Seong~Joon Oh, Sanghyuk Chun, Junsuk Choe, and Youngjoon Yoo.
\newblock {CutMix: Regularization Strategy to Train Strong Classifiers With Localizable Features}.
\newblock In {\em Proceedings of the IEEE/CVF International Conference on Computer Vision}, pages 6023--6032, 2019.

\bibitem{mixup}
Hongyi Zhang, Moustapha Cisse, Yann~N Dauphin, and David Lopez-Paz.
\newblock {Mixup: Beyond Empirical Risk Minimization}.
\newblock In {\em 6th International Conference on Learning Representations}, 2018.

\bibitem{DBLP:conf/iclr/ZhaoCDRL20}
Pu Zhao, Pin-Yu Chen, Payel Das, Karthikeyan~Natesan Ramamurthy, and Xue Lin.
\newblock {Bridging Mode Connectivity in Loss Landscapes and Adversarial Robustness}.
\newblock 2020.

\bibitem{DBLP:conf/cvpr/Zhao0XDWL22}
Zhendong Zhao, Xiaojun Chen, Yuexin Xuan, Ye Dong, Dakui Wang, and Kaitai Liang.
\newblock {DEFEAT: Deep Hidden Feature Backdoor Attacks by Imperceptible Perturbation and Latent Representation Constraints}.
\newblock In {\em Proceedings of the IEEE/CVF Conference on Computer Vision and Pattern Recognition}, pages 15213--15222, 2022.

\bibitem{imperceptible}
Nan Zhong, Zhenxing Qian, and Xinpeng Zhang.
\newblock {Imperceptible Backdoor Attack: From Input Space to Feature Representation}.
\newblock {\em arXiv preprint arXiv:2205.03190}, 2022.

\bibitem{zhou2003learning}
Dengyong Zhou, Olivier Bousquet, Thomas Lal, Jason Weston, and Bernhard Sch{\"o}lkopf.
\newblock {Learning with Local and Global Consistency}.
\newblock {\em Advances in neural information processing systems}, 16, 2003.

\end{thebibliography}
}

\clearpage
\appendix

\maketitle
\ificcvfinal\thispagestyle{empty}\fi

\section{Appendix}

\subsection{Details about Datasets}
The detailed information of the datasets used in our experiments is summarized in Table \ref{Tabel6}.

\begin{table}[htbp]
\Huge
\begin{center}
\resizebox{1\linewidth}{!}{
    \begin{tabular}{c|c|c|c|c}
    \toprule[3pt]
    Dataset & Classes & Input Size  &Training Images &Test Images \\
    \hline 
    CIFAR-10 &10 & 32 x 32 x 3 & 50000 & 10000  \\
    \hline 
    \makecell[c]{ImageNet \\ Subset} &{12} & {224 x 224 x 3} & {12480} & {2860} \\
    
    \bottomrule[3pt]
    \end{tabular}
}
\end{center}
\caption{Detailed information of the datasets used in our experiments.}
\label{Tabel6}
\end{table}

\subsection{Detailed Settings for Backdoor Attacks}
\begin{figure*}[!ht]
\begin{center}
\includegraphics[width=0.9\textwidth]{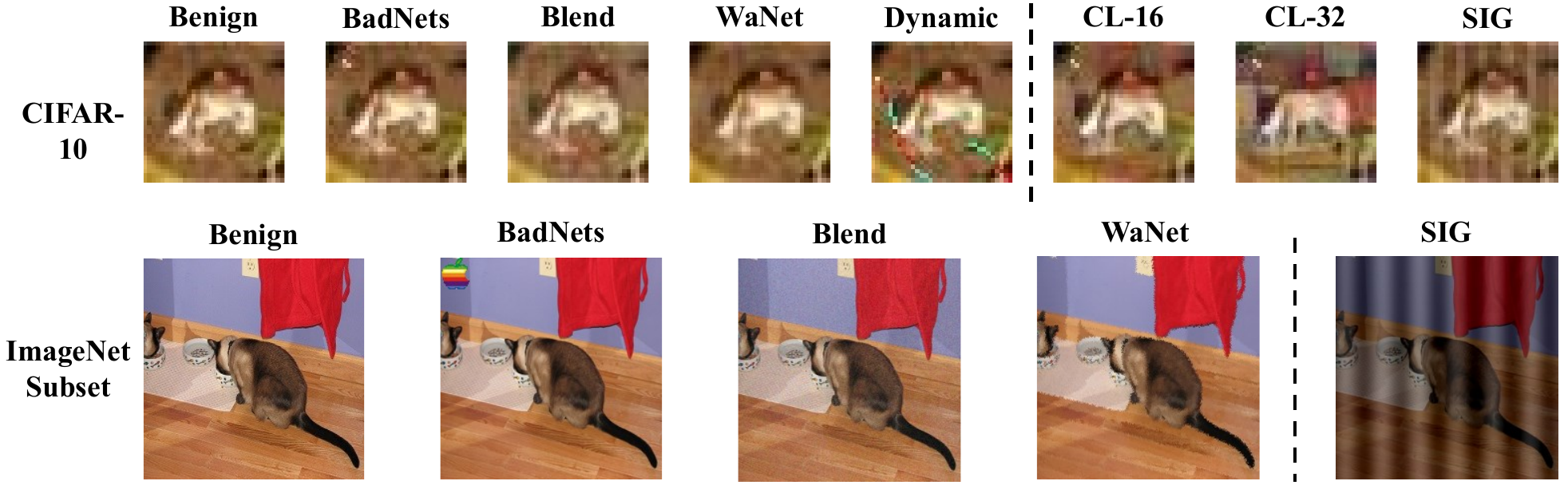}
\end{center}
   \caption{Poisoned samples crafted by different backdoor attacks for CIFAR-10 and ImageNet subset, including BadNets \cite{badnets}, Blend \cite{blend}, WaNet \cite{wanet}, Dynamic \cite{Dynamic}, CL \cite{CL} and SIG \cite{SIG}. The first sample in the two rows is benign.}
\label{fig: 6}
\end{figure*}

We trained all attack baselines for 200 epochs using the SGD optimizer with an initial learning rate of 0.1, a weight decay of 1e-4, and a momentum of 0.9. The learning rate was divided by 10 after every 50 epochs. We set the batch size to 128 for CIFAR-10 and 16 for the ImageNet subset. The poisoned samples crafted by different attacks are shown in Figure \ref{fig: 6}.

\flushleft{\textbf{Settings for BadNets \cite{badnets} }}
\justifying
We use a $2\times2$ square as the trigger on CIFAR-10 and a $32 \times 32$ Apple logo on the ImageNet subset, as suggested in previous studies \cite{DBD, neuralclense}. The triggers are added in the upper left corner of benign samples.

\flushleft{\textbf{Settings for Blend \cite{blend} }}
\justifying
We use a `Hello Kitty' image as the trigger on CIFAR-10 and a random noisy pattern on the ImageNet subset. The blended ratio is set to 0.1 for both datasets.

\flushleft{\textbf{Settings for WaNet \cite{wanet} }}
\justifying
Following the original settings, we set the grid size $k=4$ and the warping strength $s=0.5$ on CIFAR-10. But for the ImageNet subset, the grid size is set to 224 and the warping strength is set to 1 to ensure the attack works, as suggested in \cite{DBD}. We set the noisy rate $\rho_n=0.2$ for CIFAR-10 datasets and delete noisy mode for the ImageNet subset.

\flushleft{\textbf{Settings for Dynamic \cite{Dynamic} }}
\justifying
We use the pre-trained generator to generate triggers for each poisoned sample and cross-triggers for a small portion of benign samples. The cross-trigger mode rate $\rho_c$ is set to 0.1, the same as the poisoning rate.

\flushleft{\textbf{Settings for CL \cite{CL} }}
\justifying
Based on the perturbed images released by the author, we paste the BadNets trigger to them to create poisoned samples.

\flushleft{\textbf{Settings for SIG \cite{SIG} }}
\justifying
Following the original settings, we set $\triangle=20$ and $f=6$ to generate a sinusoidal signal as the trigger on CIFAR-10, and superimpose it on benign samples with a ratio of 0.1. For ImageNet, we set $\triangle=60$, $f=6$, a blend ratio of 0.5, and a poisoning rate of 0.5 to make the attack effective. 

\subsection{Warming-up Strategy Selection}

\begin{figure}[htb]
\begin{center}
\includegraphics[width=1\linewidth]{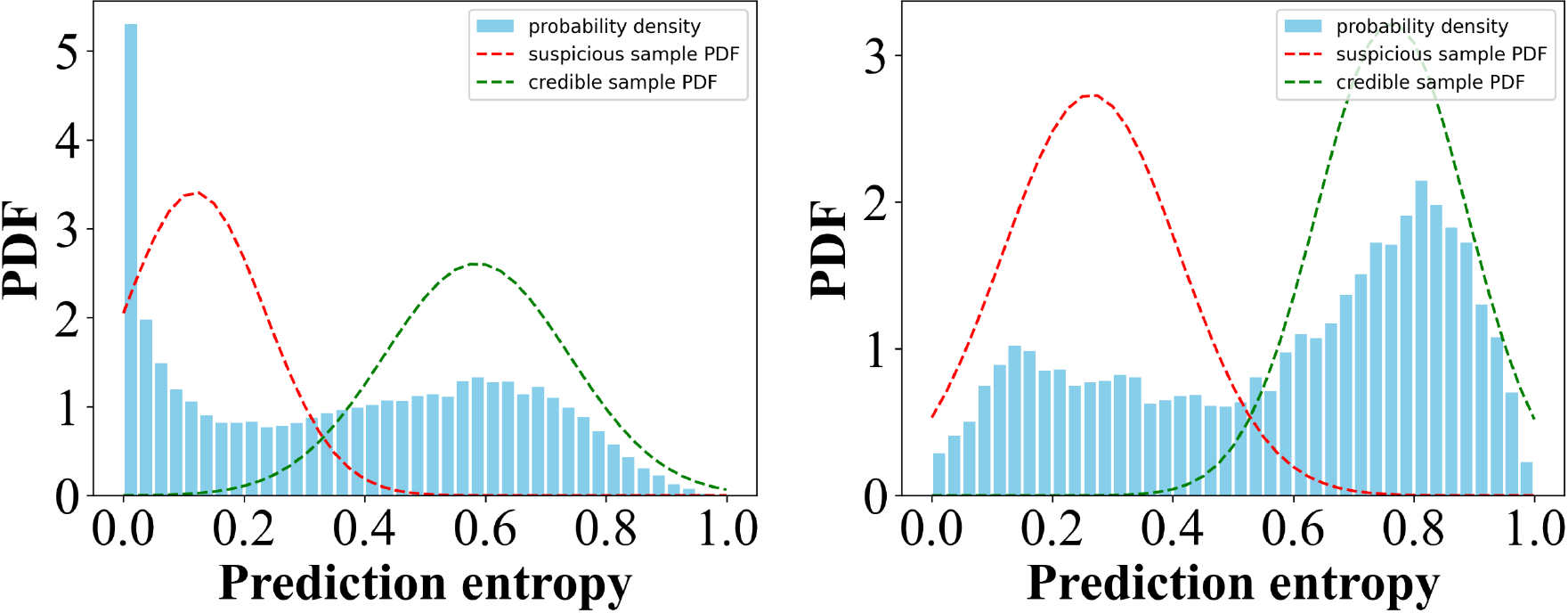}
\end{center}
   \caption{The distributions of prediction entropy on CIFAR-10 (left) and the ImageNet subset (right) under BadNets attack. The blue histogram is the probability density histogram of all samples' prediction entropy. The red and green dotted lines are two Gaussian distributions fitted by GMM, donating the distribution of suspicious samples and the distribution of credible samples, respectively.}
\label{fig: 8}
\end{figure}

In this section, we introduce the way that we select the filtering threshold $t_f$ in the warming-up stage. 

During the standard training process, the prediction entropy of poisoned samples will drop faster than benign ones, leading to two clusters in the distribution of prediction entropy. Therefore, we adopt a two-component Gaussian Mixture Model (GMM) to fit the distribution of all samples' prediction entropy using the Expectation-Maximization algorithm, as shown in Figure \ref{fig: 8}. We assume the prediction entropy of suspicious samples follows $N(\mu_1, \sigma_1^2)$ and that of benign samples follows $N(\mu_2, \sigma_2^2)$. We use the GMM to distinguish suspicious samples and train a network with them until the two distributions roughly separate, \ie $\mu_1 + 3\sigma_1 < \mu_2$. In order to retain more samples to train the Beneficiary network, we select $\mu_1 + \sigma_1$ as the filtering threshold. We conduct experiments on CIFAR-10 and ImageNet subset with BadNets and Blend attacks, finding that $\mu_1 + \sigma_1$ stops at about 0.2 after 2 to 3 epochs on CIFAR-10 and at about 0.4 after 5 to 6 epochs on the ImageNet subset. So we warm up the Victim network for 3 epochs on CIFAR-10 with $t_f$ linearly decreases from 1 to 0.2 and for 6 epochs on the ImageNet subset with $t_f$ linearly decreases from 1 to 0.4. We adopt the same settings against other attacks on both datasets and find that they still perform well. 
The reason why GMM has not been used for identification in stage 2 and stage 3 is that as the training progresses, the two distributions will gradually approach, thus affecting the ability of GMM to distinguish.

Inspired by that using GMM can well distinguish suspicious samples, we design an automatic warming-up strategy based on GMM to facilitate the selection of the filtering threshold. Similarly, we also adopt GMM to identify suspicious samples and train the Victim network with them until $\mu_1 + 3\sigma_1 < \mu_2$. Then we fix the $t_f$ to $\mu_1 + 0.5\sigma_1$ in the following training stages because we find this threshold is large enough to filter out most poisoned samples. As Tabel \ref{Tabel8} shows, the automatic warming-up strategy achieves comparable results to the original settings, which provides an alternative possibility for fast selection of the filtering threshold.

\begin{table}[htbp]
\Huge
\begin{center}
\resizebox{0.9\linewidth}{!}{
    \begin{tabular}{c|c|c|cc}
    \toprule[3pt]
    Dataset & Attack & Metric & Ours(original) & Ours(automatic) \\
    \hline
    \multirow{14}[14]{*}{CIFAR-10} & \multirow{2}[2]{*}{BadNets \cite{badnets}} & BA    & 93.96\% & \textbf{94.33\%} \\
          &       & ASR   & \textbf{0.62\%} & 0.74\% \\
\cline{2-5}          & \multirow{2}[2]{*}{Blend \cite{blend}} & BA    & \textbf{94.37\%} & 93.87\% \\
          &       & ASR   & \textbf{0.63\%} & 0.72\% \\
\cline{2-5}          & \multirow{2}[2]{*}{WaNet \cite{wanet}} & BA    & \textbf{94.15\%} & 91.74\% \\
          &       & ASR   & \textbf{0.54\%} & 1.00\% \\
\cline{2-5}          & \multirow{2}[2]{*}{CL-16 \cite{CL}} & BA    & \textbf{94.24\%} & 93.25\% \\
          &       & ASR   & 1.01\% & \textbf{0.76\%} \\
\cline{2-5}          & \multirow{2}[2]{*}{CL-32 \cite{CL}} & BA    & \textbf{93.98\%} & 93.14\% \\
          &       & ASR   & \textbf{0.64\%} & 1.17\% \\
\cline{2-5}          & \multirow{2}[2]{*}{SIG \cite{SIG}} & BA    & \textbf{94.08\%} & 93.15\% \\
          &       & ASR   & \textbf{0.17\%} & \textbf{0.01\%} \\
\cline{2-5}          & \multirow{2}[2]{*}{Dynamic \cite{Dynamic}} & BA    & \textbf{93.91\%} & 93.21\% \\
          &       & ASR   & \textbf{1.13\%} & 2.51\% \\
    \hline
    \multicolumn{1}{c|}{\multirow{8}[8]{*}{ImageNet Subset}} & \multirow{2}[2]{*}{BadNets \cite{badnets}} & BA    & 95.42\% & \textbf{95.90\%} \\
          &       & ASR   & 0.28\% & \textbf{0.24\%} \\
\cline{2-5}          & \multirow{2}[2]{*}{Blend \cite{blend}} & BA    & \textbf{95.03\%} & 92.08\% \\
          &       & ASR   & \textbf{0.45\%} & 1.54\% \\
\cline{2-5}          & \multirow{2}[2]{*}{WaNet \cite{wanet}} & BA    & \textbf{94.84\%} & 94.65\% \\
          &       & ASR   & 1.92\% & \textbf{0.49\%} \\
\cline{2-5}          & \multirow{2}[2]{*}{SIG \cite{SIG}} & BA    & 94.65\% & \textbf{95.74\%} \\
          &       & ASR   & \textbf{0.03\%} & 0.17\% \\
    \bottomrule[3pt]
    \end{tabular}%
}
\end{center}
\caption{The results of V\&B with original warming-up strategy and automatic warming-up strategy. The automatic warming-up strategy achieves comparable results to the original settings, which provides an alternative possibility for fast selection of the filtering threshold.}
\label{Tabel8}
\end{table}

\subsection{Effectiveness under Different Poisoning Rate}

\begin{figure}[h]
\begin{center}
\includegraphics[width=0.9\linewidth]{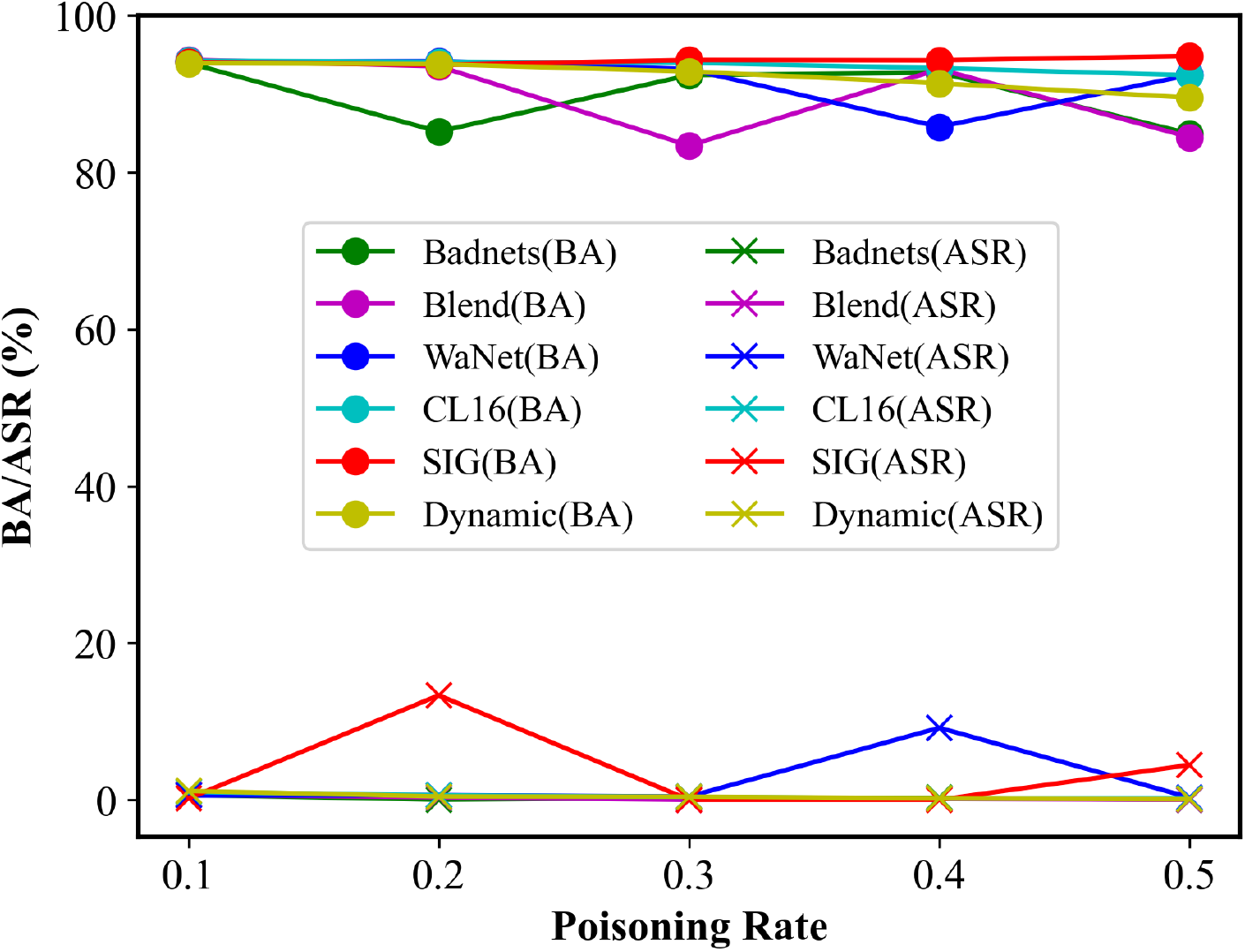}
\end{center}
   \caption{The benign accuracy and attack success rate under different poisoning rates. }
\label{fig: 7}
\end{figure}

We verified the effectiveness of our framework with poisoning rates ranging from 0.1 to 0.5, and the results are shown in Figure \ref{fig: 7}. Against most attacks, our framework can reduce the attack success rate to below 1\% at various poisoning rates, while maintaining a satisfactory benign accuracy. With the poisoning rate increasing, the Victim network can better learn trigger patterns and filter out a larger proportion of poisoned samples for the Beneficiary network. This is why our framework works fine even with a poisoning rate of 0.5. Although the proportion of filtered poisoned samples becomes larger, the number of missed poisoned samples may also increase as the poisoning rate increases. Especially when the poisoning rate is not high enough (\eg 0.2 for SIG and 0.4 for WaNet), the Victim network will miss more poisoned samples, which may cause the attack success rate to increase. 
When the poisoning rate reaches 0.5, all benign accuracy drops because fewer benign samples can be used. We can also observe that the benign accuracy of some attacks has fluctuated (\ie 0.2 for BadNets, 0.3 for Blend, and 0.4 for WaNet), which is possibly due to selecting too many benign samples of a certain class for poisoning, resulting in the model being unable to correctly relabel the poisoned samples under this class.

\subsection{Detailed Results of Our V\&B}

In Table \ref{Tabel7}, we show the results after stage 2 and stage 3 (final results) of our framework for all attack cases. It is obvious that the attack success rates after stage 2 are all higher than the final results, while the benign accuracy is the opposite. This demonstrates that our semi-supervised suppression training can both erase existing backdoors and improve model performance on benign samples.

\begin{table}[htbp]
\Huge
\begin{center}
\resizebox{0.9\linewidth}{!}{
   \begin{tabular}{c|c|c|cc}
    \toprule[3pt]
    Dataset & Attack & Metric & After stage 2 & After stage 3 \\
    \hline
    \multirow{14}[16]{*}{CIFAR-10} & \multirow{2}[2]{*}{BadNets \cite{badnets}} & BA   & 89.87\% & \textbf{93.96\%} \\
          &       & ASR   & 1.94\% & \textbf{0.62\%} \\
\cline{2-5}          & \multirow{2}[2]{*}{Blend \cite{blend}} & BA   & 90.03\% & \textbf{94.37\%} \\
          &       & ASR   & 4.94\% & \textbf{0.63\%} \\
\cline{2-5}          & \multirow{2}[2]{*}{WaNet \cite{wanet}} & BA   & 91.46\% & \textbf{94.15\%} \\          &       & ASR   & 75.33\% & \textbf{0.54\%} \\
\cline{2-5}          & \multirow{2}[2]{*}{CL-16 \cite{CL}} & BA   & 91.69\% & \textbf{94.24\%} \\
          &       & ASR   & 1.71\% & \textbf{1.01\%} \\
\cline{2-5}          & \multirow{2}[2]{*}{CL-32 \cite{CL}} & BA   & 91.09\% & \textbf{93.98\%} \\
          &       & ASR   & 1.91\% & \textbf{0.64\%} \\
\cline{2-5}          & \multirow{2}[2]{*}{SIG \cite{SIG}} & BA   & 91.27\% & \textbf{94.08\%} \\
          &       & ASR   & 3.40\% & \textbf{0.17\%} \\
\cline{2-5}          & \multirow{2}[2]{*}{Dynamic \cite{Dynamic}} & BA   & 90.03\% & \textbf{93.91\%} \\
          &       & ASR   & 76.34\% & \textbf{1.13\%} \\
    \hline
    \multicolumn{1}{c|}{\multirow{8}[8]{*}{ImageNet Subset}} & \multirow{2}[2]{*}{BadNets \cite{badnets}} & BA   & 94.42\% & \textbf{95.42\%} \\
          &       & ASR   & 0.52\% & \textbf{0.28\%} \\
\cline{2-5}          & \multirow{2}[2]{*}{Blend \cite{blend}} & BA   & 91.31\% & \textbf{95.03\%} \\
          &       & ASR   & 2.13\% & \textbf{0.45\%} \\
\cline{2-5}          & \multirow{2}[2]{*}{WaNet \cite{wanet}} & BA   & 90.87\% & \textbf{94.84\%} \\
          &       & ASR   & 4.55\% & \textbf{1.92\%} \\
\cline{2-5}          & \multirow{2}[2]{*}{SIG \cite{SIG}} & BA   & 91.76\% & \textbf{94.65\%} \\
          &       & ASR   & 2.38\% & \textbf{0.03\%} \\
    \bottomrule[3pt]
    \end{tabular}%
}
\end{center}
\caption{Detailed results of our V\&B against different attacks.}
\label{Tabel7}
\end{table}


\end{document}